%% file: neurips_2026.tex
\documentclass{article}

\usepackage[preprint]{neurips_2026}
\input{math_cmd}

\usepackage[utf8]{inputenc} %
\usepackage[T1]{fontenc}    %
\usepackage[colorlinks=true, allcolors=blue]{hyperref}       %
\usepackage{url}            %
\usepackage{booktabs}       %
\usepackage{amsfonts}       %
\usepackage{nicefrac}       %
\usepackage{microtype}      %
\usepackage{xcolor}         %
\usepackage{placeins}
\usepackage{float}

\usepackage{amsmath}
\usepackage{amssymb}

\usepackage{nameref}
\usepackage{varioref}
\usepackage{cleveref}

\usepackage{booktabs}
\usepackage{multirow}
\usepackage{makecell}
\usepackage{arydshln} %
\usepackage{graphicx} %

\usepackage[most]{tcolorbox}
\usepackage{listings}
\usepackage{titletoc}

\newtcolorbox{codebox}[1][]{
  enhanced,                %
  breakable,               %
  colback=gray!5,
  colframe=gray!75,
  fonttitle=\bfseries,
  coltitle=black,
  boxrule=0.5pt,
  arc=2mm,                 %
  outer arc=2mm,           %
  left=10pt, right=10pt,
  top=10pt, bottom=10pt,
  title=#1
}

\title{\ODRPO: Ordinal Decompositions of Discrete Rewards for Robust Policy Optimization}

\author{%
  Nirmal Patel\thanks{Work done during a Student Researcher internship at Google.} \\
  University of Texas at Austin\\
  \texttt{nirmpatel@utexas.edu} \\
  \And
  Fei Wang \\
  Google \\
  \texttt{feiwangnlp@google.com} \\
  \And
  Inderjit S. Dhillon \\
  Google \\
  \texttt{isd@google.com}
}

\begin{document}

\maketitle

\begin{abstract}
  
  The alignment of Large Language Models (LLMs) increasingly relies on Reinforcement Learning from AI Feedback (RLAIF) for non-verifiable domains such as long-form question answering and open-ended instruction following. These domains often rely on LLM based \autorater{}s to provide granular, multi-tier discrete rewards (e.g., 1-10 rubrics) that are inherently stochastic due to prompt sensitivity and sampling randomness. We empirically verify the stochasticity of \autorater{s} that can propagate and corrupt standard advantage estimators like GRPO and MaxRL, as a noisy reward samples can skew normalization statistics and degrade the global learning signal. Empirically, sampling more rewards and taking majority voting may reduce the noise and improve performance, but this approach is computationally expensive. To address this bottleneck, we introduce \ODRPOFULLBOLD~(\textbf{\ODRPO}), a framework that structurally isolates evaluation noise by decomposing discrete rewards into a sequence of ordinal binary indicators. By independently computing and accumulating advantages across these progressively challenging success thresholds, \ODRPO~prevents outlier evaluations from corrupting the global update while establishing an implicit, variance-aware learning curriculum. Empirically, \ODRPO~achieves robust performance on Qwen2.5-7B and Qwen3-4B models, frequently outperforming baselines with relative improvements of upto 14.8\% on FACTS-grounding-v2 and 7.5\% on Alpaca-Evals. Critically, these gains are achieved with negligible training-time overhead, as \ODRPO~requires no additional compute per step compared to standard estimators. Supported by theoretical analysis confirming its optimization stability, \ODRPO~provides a scalable and robust framework for aligning models within the noisy, discrete evaluation landscape of modern RLAIF. 
\end{abstract}

\section{Introduction}

Reinforcement Learning (RL) has become a prominent post-training method in the field of language modeling. With the notable success of reinforcement learning with verifiable rewards (RLVR) in fields like coding and math~\citep{Guo_2025}, there has been a push to apply RLVR to other verifiable domains, such as puzzle solving, scientific literature review, and visual perception~\citep{chen2025enigmatascalinglogicalreasoning, burgess2026papersearchqalearningsearchreason, wang2025vicritverifiablereinforcementlearning}. However, many tasks are inherently non-verifiable, such as long-form question answering and open-ended instruction following, as they lack a fixed ground-truth and cannot be trivially addressed by RLVR~\citep{liu2026examiningreasoningllmsasjudgesnonverifiable, gunjal2026rubrics}. For these open-ended domains, \autorater{}s or ``LLM-as-judge'' frameworks are preferred due to their rapid annotation capabilities and flexible rubric adherence, acting as surrogate reward functions that align closely with human preferences~\citep{lee2024rlaif, gunjal2026rubrics}. \Autorater{}s also offer the flexibility to output discrete rewards based on detailed rubrics or partial scoring, providing more granular learning signals~\citep{kwok2026llmverifier}. Consequently, reinforcement learning from AI feedback (RLAIF) has emerged as a scalable and reliable post-training paradigm for non-verifiable domains~\citep{lee2024rlaif, liu2026examiningreasoningllmsasjudgesnonverifiable, gunjal2026rubrics}. 

While RLAIF is becoming a standard practice in non-verifiable domains, it introduces the challenge of noisy reward signals driven by the inherent stochasticity of the \autorater. This noise can stem from prompt sensitivity, position bias, and rubric misinterpretation~\citep{zhao2025tokenfoolllmasajudge, shi2025judgingjudgessystematicstudy, li2026evaluatingscoringbiasllmasajudge}. 
This unpredictability poses a significant challenge for standard advantage estimators like GRPO~\citep{shao2024deepseekmathpushinglimitsmathematical} and MaxRL~\citep{tajwar2026maximumlikelihoodreinforcementlearning}, which implicitly assume reliability of the reward signal. Under the stochasticity of the \autorater, the advantage estimates can become corrupted. A single noisy reward sample can skew the normalization statistics, negatively impacting the entire group's update. While Monte-Carlo sampling could mitigate this variance, it remains computationally expensive and intractable at scale, as it multiplies the cost of \autorater calls, which are often more expensive than the response sampling from the policy model. 

In this work, we propose \ODRPOFULLBOLD~(\ODRPO), a framework designed to stabilize the optimization signal in RLAIF within discrete reward settings. We decompose the single scalar reward into multiple sub-rewards representing ordinal success levels, compute the advantage for each level independently, and then accumulate these values, effectively transforming a single-reward optimization into a multi-reward formulation. Decomposing the reward into ordinal thresholds confines evaluator stochasticity to narrow boundary strata, preventing a single noisy evaluation from corrupting the global mean and variance used for normalization, and preserving a stable learning signal across the rest of the reward distribution. Furthermore, we introduce variance-aware weighting schemes, utilizing dynamic weighting functions that systematically suppress the noise originating from completely solved or currently unreachable evaluation criteria, focusing the model's capacity on active learning frontiers. Crucially, this structural decomposition introduces no measurable training-time overhead, making \ODRPO~a highly efficient ``plug-and-play'' augmentation for traditional advantage estimators in large-scale alignment workflows. 

We evaluated \ODRPO~in the RLAIF setting against standard GRPO and MaxRL estimators on three diverse benchmarks: FACTS-grounding-v2~\citep{cheng2025factsleaderboardcomprehensivebenchmark}, Alpaca-Eval~\citep{alpaca_eval}, and IFEval~\citep{zhou2023instructionfollowingevaluationlargelanguage}. Our empirical results demonstrate that \ODRPO~consistently improves upon baselines, achieving relative performance gains of upto 14.8\% on FACTS-grounding-v2 and 7.5\% on Alpaca-Evals. These gains highlight the framework's ability to extract robust alignment signals from stochastic evaluators. Alongside these empirical relative improvements, we provide a theoretical analysis illustrating how ordinal decomposition ensures the estimator admits a well-defined global scalar objective, thereby stabilizing the optimization trajectory. 

Our core contributions are threefold. First, we empirically identify the stochasticity of \autorater{}s in multi-tier discrete reward space and analyze its variance and effects on rank-flips. Second, we introduce \ODRPO, a novel advantage estimation framework featuring ordinal reward decomposition and Variance-Aware Weighting Schemes to structurally isolate and suppress evaluator noise. Third, we provide comprehensive empirical validation demonstrating robust improvements across diverse open-ended alignment tasks, supported by theoretical analysis confirming the framework's stability.

\section{Related Work}\label{sec:RelatedWork}

The alignment of Large Language Models (LLMs) with human intent and logical consistency has primarily consolidated around the framework of Reinforcement Learning from Human Feedback~(RLHF). Early advancements were largely catalyzed by Proximal Policy Optimization (PPO)~\citep{schulman2017proximalpolicyoptimizationalgorithms}, which relies on a centralized critic network to mitigate gradient variance. While PPO remains a robust baseline, the substantial computational overhead required to maintain auxiliary value models has motivated the search for more resource-efficient alternatives. A significant development in this direction is Group Relative Policy Optimization (GRPO)~\citep{shao2024deepseekmathpushinglimitsmathematical}, which avoids the traditional critic in favor of computing advantages relative to a group of sampled outputs. By utilizing these local surrogates, GRPO significantly reduces memory footprint while maintaining high performance on complex mathematical reasoning tasks. 

Concurrently, Direct Preference Optimization (DPO)~\citep{rafailov2024directpreferenceoptimizationlanguage} has emerged as a prominent offline alternative, reformulating the alignment task into a supervised cross-entropy loss. While DPO provides theoretical clarity for preference-based data, it is inherently limited by the diversity of static datasets and lacks the dynamic exploration capabilities required for reasoning-intensive domains~\citep{mohammadi2025evaluatinggrpodpofaithful}. For these tasks, online reinforcement learning continues to be the preferred paradigm. 
To further scale these methods, Reinforcement Learning from AI Feedback (RLAIF)~\citep{lee2024rlaif} utilizes high-capacity models as proxy evaluators, providing a scalable mechanism for generating synthetic reward signals. This shift toward automated, multi-faceted feedback further necessitates a rigorous theoretical understanding of the objectives that online estimators actually optimize. 

Recent theoretical advancements have formalized the global optimization landscape of policy gradient methods, demonstrating that prominent advantage estimators such as GRPO, Rejection Sampling, and REINFORCE admit well-defined global scalar objectives in binary reward settings~\citep{davis2025objectivereasoningreinforcementlearning}. Building progressively upon this framework, MaxRL delineates a path to invoke maximum-likelihood behavior by analyzing the theoretical connection between rejection sampling and the Maclaurin series expansion of the logarithm, establishing that the estimator converges to a true maximum-likelihood objective as the group size increases~\citep{tajwar2026maximumlikelihoodreinforcementlearning}. This theoretical convergence is supported by experiments demonstrating that MaxRL achieves performance closely matching standard cross-entropy baselines.%
Tangentially, GDPO identifies a structural limitation in GRPO when applied to multiple binary reward environments, where bundling distinct reward tuples into a single advantage value inherently reduces signal expressiveness~\citep{liu2026gdpogrouprewarddecouplednormalization}. To address this, GDPO decouples the advantage calculation for each individual reward prior to accumulation, thereby preserving a denser and more informative optimization signal than standard GRPO~\citep{liu2026gdpogrouprewarddecouplednormalization}. 

\ODRPO~synthesizes and extends these foundational principles to address the unique challenges of arbitrary discrete reward spaces. While previous analyses established scalar objectives for binary outcomes~\citep{davis2025objectivereasoningreinforcementlearning}, extending these estimators to arbitrary discrete rewards introduces update field asymmetries that make the scalar objectives theoretically inadmissible. Consequently, standard estimators like GRPO and MaxRL fail to admit global scalar objectives in this context. \ODRPO~resolves this theoretical gap by providing a formal framework that guarantees estimators with valid binary objectives retain a global scalar objective in the discrete reward setting. Furthermore, while \ODRPO leverages decoupled advantage calculations akin to GDPO, it applies this mechanism to create ordinal levels within a single discrete reward scale rather than parallel binary tasks. In this single discrete reward context, GDPO remains equivalent to GRPO and thus suffers from the same theoretical limitations. This success-ordinal decomposition establishes an implicit curriculum~\citep{narvekar2020curriculumlearningreinforcementlearning}, structurally guiding the policy optimization process across progressively challenging quality thresholds. 

\section{Auto-rater Rewards Exhibit High Variance} \label{sec:Problem}  %

RLAIF has become increasingly important and widely used when extending from RLVR to non-verifiable domains. However, evidence from recent studies suggests that LLM-based \autorater{}s often exhibit high variance due to prompt sensitivity and sampling randomness~\citep{zhao2025tokenfoolllmasajudge, shi2025judgingjudgessystematicstudy, li2026evaluatingscoringbiasllmasajudge}. This raises concerns about whether such stochasticity manifests as inconsistent reward signals, which would fundamentally undermine their reliability as stable feedback for RL. 

An \autorater is characterized as stochastic if $N$ independent evaluations of $M$ responses to a fixed prompt yield frequent rank-flips, manifesting as a lack of consensus and high evaluative variance. Rank agreement can be quantified using Kendall's coefficient of concordance, or Kendall's $W$~\citep{kendallW}. If $W\sim0$ then the \autorater is highly inconsistent and if $W\sim1$ then the \autorater is consistent. We conjecture that $W\geq0.9$ indicates that the \autorater is fairly consistent for stable post-training as the $N$ ``virtual judges'' are derived from the same \autorater. 

\begin{figure}[t]
  \centering
  \includegraphics[width=0.7\textwidth]{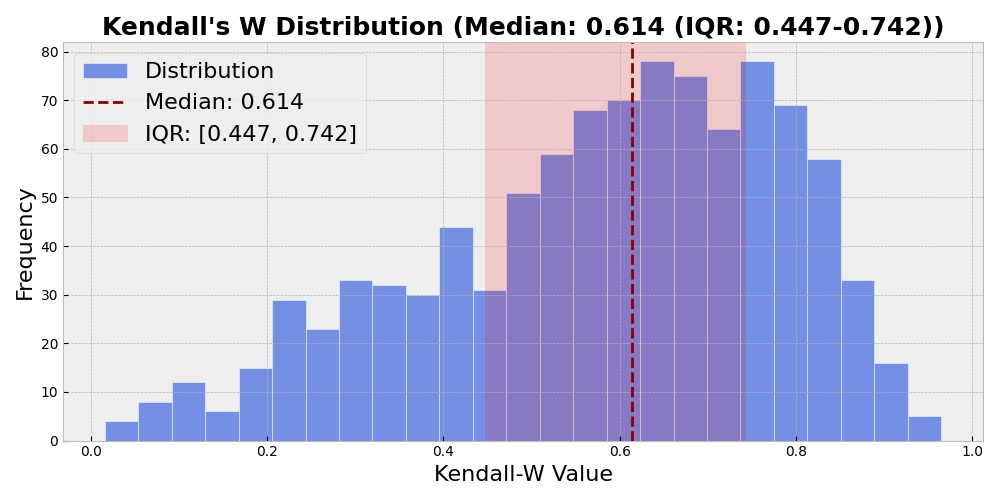}
  \caption{Kendall's coefficient of concordance for 1,000 datapoints from the Ultrafeedback dataset~\citep{cui2023ultrafeedback} for Qwen3-30B-A3B-Instruct-2507~\citep{qwen3technicalreport} as \autorater and Qwen2.5-7B-Instruct~\citep{qwen2} as the response generator. A median Kendall's W of $0.614$ with a heavy left tail indicates that the \autorater has noticeable stochasticity, making its judgment less reliable. }\label{fig:Kendall}
\end{figure}

We performed a statistical analysis using Qwen3-30B-A3B-Instruct-2507~\citep{qwen3technicalreport} as \autorater and Qwen2.5-7B-Instruct~\citep{qwen2} as the response generator on 1000 randomly sampled datapoints from the Ultrafeedback dataset~\citep{cui2023ultrafeedback}. This sample of 1000 datapoints is representative because the Ultrafeedback dataset is inherently broad and rich~\citep{jiang2025hummerlimitedcompetitivepreference, deng2026moreimprovingllmalignment}. For each datapoint, we generated $M=8$ responses and for each response we extracted $N=16$ scores between 1-10; this gives 1000 $M\times{}N$ matrices. We performed the test of significance of Kendall's coefficient of concordance (Kendall's W) on the 1000 $M\times{}N$ matrices through Pingouin~\citep{Vallat2018}. The Kendall's W value for different datapoints is shown in~\cref{fig:Kendall}, and further analysis is provided in~\cref{appsec:AutoraterAnalysis}. 

This empirical assessment demonstrates that high-capacity models like Qwen3 can produce inconsistent scoring distributions for identical inputs, resulting in a high-variance reward signal. This stochasticity propagates through the advantage estimator, potentially corrupting the gradient signal and destabilizing the optimization process. These findings highlight the necessity of a variance-aware advantage estimator--one capable of dissipating evaluative noise across ranking levels to localize and mitigate its impact on policy updates. 

\section{\ODRPO: Ordinal Decomposition for Robust Policy Optimization }

{ %
\renewcommand{\arraystretch}{2.2} %
\setlength\extrarowheight{0.1pt}  %
\setlength\dashlinedash{2pt}      %
\setlength\dashlinegap{2pt}       %
\ADLdrawingmode{2}                %

\begin{table}[ht]
  \centering
  \small
  
  \begin{tabular}{lll}
    \Xhline{1.5pt}
    \makecell{\textbf{Advantage} \\ \textbf{Estimator}} & \textbf{Formula} & \textbf{Description} \\
    \Xhline{1.2pt}

    GRPO & $A_i = \frac{r_i - \mu}{\sigma}$ & Z-Score normalization of the rewards \\
    \hdashline

    MaxRL & $A_i = \frac{r_i - \mu}{\mu}$ & \makecell[l]{Normalization by mean to approach \\ maximum-likelihood} \\
    \hdashline

    \ODRPO~& \makecell[l]{$A_i = \sum_{k=1}^K A_i^{(k)}$ where $A_i^{(k)} = \frac{r_i^{(k)} - \mu^{(k)}}{N^{(k)}}$,\\$r_i^{(k)} = \mathbb{I}\{r_i \geq k\}$} & \makecell[l]{For a discrete integer reward in range \\ $\{1, \dots, K\}$, first compute advantage \\ of each ordinal level and then \\ accumulate} \\
    \Xhline{1.5pt}
  \end{tabular}
  \vspace{3pt}
  \caption{Formulae and description of different advantage estimators. $r_i$ is the $i^\text{th}$ rollout's reward and $A_i$ is its advantage value. $\mu$ and $\sigma$ are the group's mean and standard deviation while $\mu^{(k)}, \sigma^{(k)}, N^{(k)}$ are the $k^\text{th}$ level's mean, standard deviation, and normalization, respectively. For GRPO-like normalization $N^{(k)}=\sigma^{(k)}$ and for MaxRL-like normalization $N^{(K)}=\mu^{(k)}$.} \label{tab:Formulas}
\end{table}
}

To address the instability caused by \autorater stochasticity in multi-tier reward spaces, we propose the \ODRPOFULLBOLD~(\ODRPO) framework. Instead of treating a discrete rubric score as a single continuous value, \ODRPO restructures the objective by decomposing the reward into a monotonic sequence of binary sub-tasks. This transformation isolates evaluation noise and establishes a stable, variance-aware learning trajectory. The formulations of ODRPO along with GRPO and MaxRL and their descriptions are provided in~\cref{tab:Formulas}. 

\subsection{Ordinal Decomposition}\label{subsec:OrdinalDecomposition}

To align models with multi-tier evaluation signals, the advantage estimator must accommodate the discrete structure of reward rubrics. Instead of aggregating these signals into a single cardinal value, we propose a success-ordinal decomposition that treats rewards as a sequence of progressive success thresholds. Given a discrete reward space $\mathcal{R}:=\{1, 2, \ldots, K\}$ within a group $\mathcal{G}$, we decompose the observed reward $r_i$ into a set of binary indicators $r_i^{(k)}$, denoting whether the $i^\text{th}$ rollout meets the $k^\text{th}$ quality level. By treating the discrete reward space as an ordered sequence of sub-tasks rather than a monolithic continuous variable, this decomposition preserves the granularity of the evaluation signal. 

\ODRPO~applies decoupled normalization to the ordinal levels of a single discrete scale and then aggregates. We compute the advantage for each ordinal bin independently and accumulate them to yield the final advantage: 
\begin{align*}
  r_i &= \sum_{k=1}^K r_i^{(k)},\quad r_i^{(k)} = \mathbb{I}\{r_i \geq k\}, \\
  A_i^{(k)} &= \frac{r_i^{(k)} - \mu^{(k)}}{N^{(k)}}, \\
  A_i &= \sum_{k=1}^K A_i^{(k)}, 
\end{align*}

where $\mu^{(k)} = \mathbb{E}_{i\in \mathcal{G}}[r_i^{(k)}]$ is the mean of the $k^\text{th}$ bin, and $N^{(k)}$ is a bin-specific normalization factor (e.g., standard deviation for GRPO or mean for MaxRL). This independent normalization across bins ensures that the variance from highly concentrated reward distributions at lower thresholds does not improperly scale the advantage computation at sparse, higher-tier boundaries. 

By partitioning the reward space into success indicators, \ODRPO~maintains the monotonic progression of evaluation rubrics. This formulation introduces an implicit curriculum~\citep{narvekar2020curriculumlearningreinforcementlearning}, structuring the policy update to stabilize fundamental quality thresholds before optimizing for higher success levels. Moreover, by structurally isolating distinct performance boundaries, \ODRPO~prevents optimization gradients from being dominated by outlier responses or trivial successes. This localized advantage calculation directly translates dense qualitative rubrics into actionable, bounded updates. 

\subsection{Variance-Aware Weighting Schemes}\label{subsec:WeightsOfOrdinal}

The success-ordinal framework permits weighting the optimization signal across different reward thresholds. While linear weighting can emphasize higher-tier rewards, we present two variance-aware schemes designed to focus updates on the most informative thresholds: (i) \GINI~weighting and (ii) \GINIMEDIAN~weighting. These dynamic weighting functions systematically suppress the noise originating from completely solved or currently unreachable evaluation criteria. 

The \GINI~weighting scheme uses scaled \GINI~impurity, $4\mu^{(k)} (1 - \mu^{(k)})$, to assign higher weights to bins with a mean reward near 0.5. This emphasizes thresholds that are neither trivially satisfied nor overly difficult under the current policy. We incorporate a $\sqrt{k}$ factor to maintain optimization pressure toward higher quality: 
\begin{align}\label{eq:GiniWeightings}
  w^{(k)}_\text{\GINI} &= \sqrt{k} \cdot (0.1 + 4\mu^{(k)} (1 - \mu^{(k)})).
\end{align}

The \GINIMEDIAN~scheme modulates the \GINI-impurity using a biased exponential decay relative to the group's median bin index $M_\mathcal{G}$. This directs the optimization toward the median-to-higher tiers, anchoring updates to the model's current performance: 
\begin{align}\label{eq:GiniMedianWeightings}
  w^{(k)}_\text{\GINIMED} &= \sqrt{k} \cdot \left(0.1 + 4\mu^{(k)} (1 - \mu^{(k)})\cdot \exp\left( -\frac{\max(M_\mathcal{G} - k, 0)}{2} \right) \right).
\end{align}

\begin{figure}[ht]
  \centering
  \includegraphics[width=0.9\textwidth]{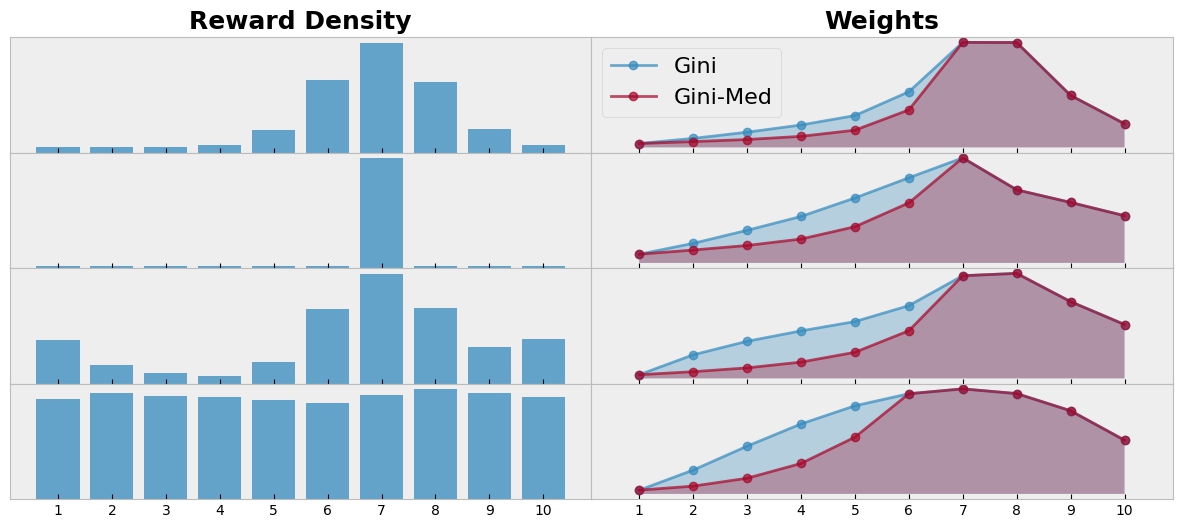}
  \caption{Visualization of \GINI\ and \GINIMED\ weighting behaviors across four representative reward distributions: (a) standard normal, (b) concentrated peak, (c) normal with bi-directional outliers, and (d) uniform. While both variants dynamically adapt to the reward density to prioritize the learning frontier, \GINIMED\ attenuates the weighting of reward levels below the median. This modulation concentrates the optimization signal on the frontier and high-difficulty levels, minimizing the influence of already resolved or saturated reward states on the policy update.}\label{fig:WeightingAnalysis}\label{fig:GiniWeights}
\end{figure}

The accumulated advantage is then computed as $A_i = \sum_{k=1}^K w^{(k)} A_i^{(k)}$, where $w^{(k)} = 1$ recovers the unweighted aggregation. As illustrated in~\cref{fig:GiniWeights}, these schemes allow \ODRPO~to adapt the optimization landscape based on the empirical reward distribution. By concentrating learning capacity on active performance frontiers, the variance-aware mechanisms accelerate convergence while preventing catastrophic forgetting of stabilized lower-tier behaviors. Consequently, the weighting strategies minimize the sample inefficiency associated with uniform reward distribution assumptions. They provide a statistically principled method for adjusting the optimization target as the policy's competency shifts across the ordinal spectrum. 

The accumulated advantage will be computed as $A_i = \sum_{k=1}^K w^{(k)} A_i^{(k)}$, where $w^{(k)}$ are the weights assigned by the chosen weighting scheme. The flexibility in weighting allows \ODRPO~to adapt to various reward structures and task requirements, making it a versatile framework for robust advantage estimation in preference finetuning. 

\section{Experiments}\label{sec:Experiments}

\textbf{Setup. }We trained Qwen2.5-7B-Instruct~\citep{qwen2} on the Ultrafeedback dataset~\citep{cui2023ultrafeedback} using the VERL framework~\citep{sheng2024hybridflowVERL}. Qwen3-30B-A3B-Instruct-2507~\citep{qwen3technicalreport} served as the \autorater, providing integer rewards (1--10). To evaluate the robustness of our framework, we compared it against GRPO~\citep{shao2024deepseekmathpushinglimitsmathematical} and MaxRL~\citep{tajwar2026maximumlikelihoodreinforcementlearning} estimators. We used batch normalization after advantage accumulation for numerical stability~\citep{liu2026gdpogrouprewarddecouplednormalization}. All models were trained for one epoch with a batch size of 64 and 8 rollouts per group. Detailed hyperparameters and prompt templates are provided in Appendix~\cref{appsec:ExpDetails} and selected training curves and their analysis is provided in Appendix~\cref{appsec:TrainingAnalysis}. 

\subsection{Evaluation Suite}\label{subsec:EvaluationSuite}

We evaluated the models on three broad benchmarks: (i) FACTS-grounding-v2~\citep{cheng2025factsleaderboardcomprehensivebenchmark}, (ii) Alpaca-Evals~\citep{alpaca_eval}, and (iii) IFEval~\citep{zhou2023instructionfollowingevaluationlargelanguage}. 

\textbf{FACTS-grounding-v2 (FACTS). }FACTS-grounding-v2 evaluates the factual grounding of the models by providing user query and context document (upto 32K tokens) and uses an ensemble of LLM judges to evaluate the response (upto 1024 tokens) by providing a binary score~\citep{cheng2025factsleaderboardcomprehensivebenchmark}. We used Gemini 3.1 Flash-Lite~\citep{gemini31flashlite2026} as a quality filtering judge. The filtered responses were then evaluated by Gemini 3 Flash~\citep{gemini3flash2025} and Gemini 2.5 Flash~\citep{gemini25flash2025} and the final score is the mean of the two judges. The responses that failed the quality filter were assigned a score of 0. All models' temperatures were set to 0. We report the overall score of each model on this benchmark. 

\textbf{Alpaca-Evals. }Alpaca-Evals is an automatic, LLM-based evaluator that evaluates models on 805 datapoints against a set baseline, which in our case is the policy model before training on the Ultrafeedback dataset~\citep{alpaca_eval}. We have Gemini 3 Flash~\citep{gemini3flash2025} as the LLM judge that chooses between two given responses and provides a ternary score: win, lose, or tie. All models' temperatures were set to 0. We report the overall length-controlled win rate of each model against the baseline.

\textbf{IFEval. }IFEval benchmarks models on instruction following capabilities through 25 types of instructions~\citep{zhou2023instructionfollowingevaluationlargelanguage}. We evaluated the models on the IFEval through the LM Eval Harness~\citep{evalHarness}, which provides a standardized interface for evaluating language models across a wide range of tasks, and report the average of the strict and loose accuracies of the models on the prompt and instruction levels. 

\subsection{Results}\label{subsec:Results}

The results of individual benchmarks and their averages are provided in~\cref{tab:Results}. The win-rates for Alpaca-Evals are calculated against the pre-finetuned policy models. Based on these data, we identify several key observations regarding the performance, robustness, and efficiency of the \ODRPO~framework.

\begin{table}[t]
  \centering
  \footnotesize 
  \setlength{\tabcolsep}{5pt} 

  \begin{tabular}{l c c c c c}
    \toprule
    & \textbf{FACTS-grounding-v2} & \textbf{Alpaca-Evals} & \textbf{IFEval} & \textbf{Mean} & \boldmath$\Delta(\%)$ \\
    \midrule
    \multicolumn{6}{l}{\textbf{Qwen2.5-7B-Instruct GRPO}} \\
    \midrule
    GRPO                & 0.1904 & 0.5691 & 0.6478 & 0.4691 & 0 \\
    \hdashline\noalign{\smallskip}
    \ODRPO~\GINI~(GRPO)        & \textbf{0.2009} & 0.5711 & 0.6526 & 0.4749 & 1.2364 \\
    \ODRPO~\GINIMED~(GRPO)    & 0.1998 & \textbf{0.5812} & \textbf{0.6611} & \textbf{0.4807} & \textbf{2.4728} \\
    \midrule
    \multicolumn{6}{l}{\textbf{Qwen2.5-7B-Instruct MaxRL}} \\
    \midrule
    MaxRL                   & 0.1419 & 0.5323 & \textbf{0.6612} & 0.4451 & 0 \\
    \hdashline\noalign{\smallskip}
    \ODRPO~\GINI~(MaxRL)      & 0.1595 & 0.5691 & 0.6451 & 0.4579 & 2.8758 \\
    \ODRPO~\GINIMED~(MaxRL)  & \textbf{0.1629} & \textbf{0.5721} & 0.6441 & \textbf{0.4597} & \textbf{3.2802} \\
    \midrule
    \multicolumn{6}{l}{\textbf{Qwen3-4B-Instruct GRPO}} \\
    \midrule
    GRPO                & 0.1215 & 0.5529 & \textbf{0.6519} & 0.4421 & 0 \\
    \hdashline\noalign{\smallskip}
    \ODRPO~\GINI~(GRPO)        & \textbf{0.1302} & \textbf{0.5661} & 0.6468 & 0.4477 & 1.2667 \\
    \ODRPO~\GINIMED~(GRPO)    & 0.1297 & 0.5641 & \textbf{0.6519} & \textbf{0.4486} & \textbf{1.4703} \\
    \midrule
    \multicolumn{6}{l}{\textbf{Qwen3-4B-Instruct MaxRL}} \\
    \midrule
    MaxRL                   & 0.1174 & 0.5533 & 0.6429 & 0.4379 & 0 \\
    \hdashline\noalign{\smallskip}
    \ODRPO~\GINI~(MaxRL)      & 0.1274 & \textbf{0.5634} & 0.6541 & 0.4483 & 2.3750 \\
    \ODRPO~\GINIMED~(MaxRL)  & \textbf{0.1332} & 0.5597 & \textbf{0.6644} & \textbf{0.4524} & \textbf{3.3113} \\
    \bottomrule
  \end{tabular}
  \vspace{3pt}
  \caption{Comprehensive benchmark results for Qwen2.5-7B and Qwen3-4B policy models using standard and \ODRPO-enhanced estimators across various alignment tasks. Evaluation metrics include factual grounding (FACTS-grounding-v2), conversational alignment (Alpaca-Evals), and instruction following (IFEval). \ODRPO~variants (\GINI~and \GINIMED) consistently outperform the base GRPO and MaxRL estimators across most benchmarks, with relative improvements up to $3.5\%$. Bold values denote the best performance in each column for each model-estimator group.}\label{tab:Results}
\end{table}

\textbf{Effectiveness Across Diverse Evaluation Benchmarks.} 
The \ODRPO~variants consistently outperform baseline estimators across all three evaluation dimensions: factual grounding, conversational alignment, and instruction following. In the Qwen2.5-7B GRPO configuration, \ODRPO~\GINIMED~improves the FACTS-grounding-v2 score from 0.1904 to 0.1998 and the Alpaca-Evals win-rate from 0.5691 to 0.5812. A similar trend is observed in the Qwen3-4B MaxRL setup, where IFEval performance increases from 0.6429 to 0.6644. These results indicate that by treating discrete rewards as ordinal levels rather than scalar, \ODRPO~effectively enhances the optimization signal. This ordinal approach prevents the advantage estimator from being skewed by outlier reward values, which is particularly beneficial for precision-critical tasks like factual grounding. 

\textbf{Consistency Across Underlying Advantage Estimators.} 
The performance gains provided by \ODRPO~are agnostic to the underlying reinforcement learning objective, showing improvements in both group-relative (GRPO) and maximum-likelihood (MaxRL) settings. For the Qwen2.5-7B model, \ODRPO~\GINIMED~yields a relative improvement ($\Delta$) of 2.47\% over the GRPO baseline and 3.28\% over the MaxRL baseline. This consistency suggests that the success-ordinal framework is not tied to a specific loss function but rather improves the fundamental quality of the advantage signal. By providing a more reliable gradient direction across different algorithmic constraints, \ODRPO~serves as a robust drop-in replacement for standard cardinal estimators. 

\textbf{Generalizability Across Model Capacities.} 
The framework demonstrates scalability across different architecture sizes, with consistent gains observed for both the 7B and 4B models. In the MaxRL configuration, \ODRPO~\GINIMED~achieves a relative gain of 3.28\% for the Qwen2.5-7B model and 3.31\% for the Qwen3-4B model. In the GRPO settings, relative improvements range from 1.23\% to 2.47\% across both scales. These data points confirm that the benefits of ordinal partitioning are not architecture-dependent. The ability to maintain similar relative performance increases in smaller capacity models (4B) suggests that \ODRPO~is a viable strategy for optimizing models where the representational capacity for complex reward signals may be more limited. 

\textbf{Computational Efficiency and Training Stability.} 
Empirical measurements confirm that the structural advantages of \ODRPO~are achieved without introducing significant computational overhead. For instance, in the Qwen2.5-7B GRPO setting, \ODRPO~\GINIMED~requires $30.37 \pm 1.17$\,s per step, compared to the baseline's $31.14 \pm 2.00$\,s. Similar parity is observed in the Qwen3-4B MaxRL configuration ($36.04 \pm 1.69$\,s vs. $36.26 \pm 1.78$\,s). These results indicate that the framework extracts a higher-fidelity signal from single-sample rollouts without increasing the training-time compute budget. This efficiency makes \ODRPO~a resource-efficient alternative to computationally expensive oversampling methods, providing a stable optimization trajectory through better structural modeling of the reward space rather than increased sample volume. 

\textbf{Takeaway.} 
The experimental evidence indicates that representing rewards via ordinal levels offers a robust alternative to traditional cardinal advantage estimation. \ODRPO~variants yield aggregated relative improvements of up to 3.3\% across diverse model capacities and advantage estimators. By strategically discarding the assumptions of a linear reward scale, the framework provides a precise and enhances optimization objective that aligns with the fundamentally discrete nature of LLM reward signals. 

\subsection{Majority Voting Ensemble Analysis}\label{subsec:MajorityVotingEnsembleAnalysis}

The inherent stochasticity of \autorater{}s can be suppressed by oversampling rewards per rollout and then taking majority voting (mode value). To analyze \ODRPO's performance against MaxRL baseline in reward oversampling regime, we conducted post-training with $N = 1, 8, 16, 32$ rewards per rollout with Qwen2.5-7B-Instruct~\citep{qwen2} policy model, Ultrafeedback dataset~\citep{cui2023ultrafeedback}, and Qwen3-30B-A3B-Instruct-2507~\citep{qwen3technicalreport} \autorater.

\begin{figure}[t]
  \centering
  \includegraphics[width=0.9\textwidth]{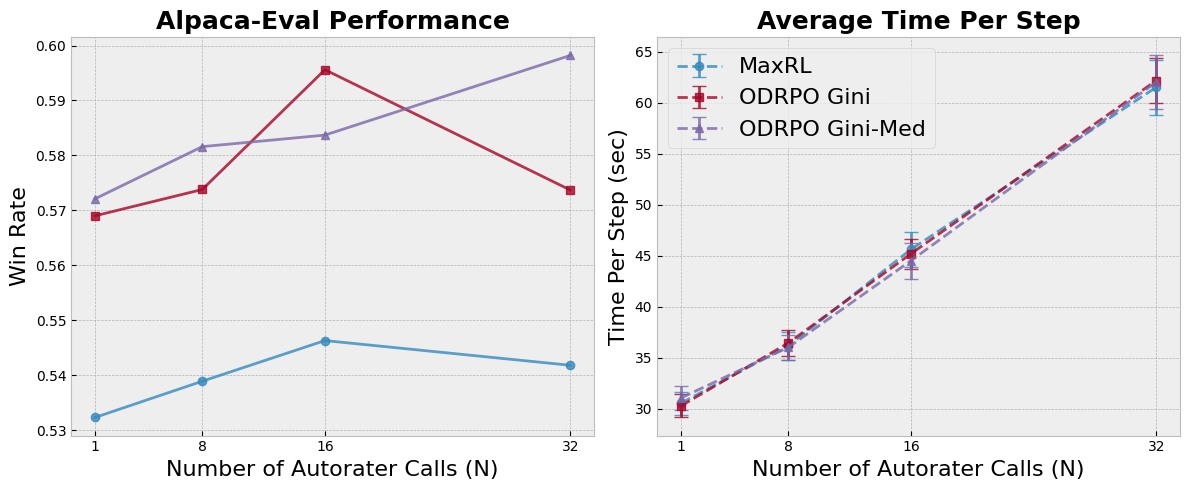}
  \caption{Alpaca-Evals values and time per step in seconds for majority voting ensemble analysis. $N=1,8,16,32$ \autorater calls were made to sample numeric rewards between 1--10 and then majority voting (mode) was used as the final reward for the corresponding rollout. \ODRPO~variants are maintaining the improvement gap against the baseline at no additional compute time. }\label{fig:FinalRewardsMode}
\end{figure}

The Alpaca-Eval values and time per step (seconds) for MaxRL against \ODRPO~methods are shown in~\cref{fig:FinalRewardsMode}. The analysis concludes that \ODRPO~maintains the improvement gap against the baseline across all oversampling regimes while adding no additional training overhead. Interestingly, $N=1$ \ODRPO~variants are observed to perform better than the best over-sampled baseline~(\GINI's 0.5690 and \GINIMED's 0.5721 against $N=16$ baseline's 0.5463). Furthermore, while $N=32$ MaxRL and \ODRPO~\GINI~MaxRL experienced a slight dip in performance, \ODRPO~\GINIMED~MaxRL continued to improve, reaching a score of 0.5982. This suggests a higher performance ceiling for \ODRPO~variants ($\sim$0.60) compared to MaxRL ($\sim$0.55), further attesting to \ODRPO's capability in decomposing rewards and localizing \autorater noise. The corresponding training analysis is provided in~\cref{appsec:MajorityVotingTraining}. 

\section{Theoretical Analysis}

Extending policy gradient methods to discrete reward spaces requires verifying that the advantage estimator optimizes a global scalar objective. While previous work identifies these objectives for binary rewards~\citep{davis2025objectivereasoningreinforcementlearning}, our analysis (\cref{appsubsec:CurlCondition}) indicates that applying estimators such as GRPO~\citep{shao2024deepseekmathpushinglimitsmathematical} or MaxRL~\citep{tajwar2026maximumlikelihoodreinforcementlearning} directly to multi-tier discrete rewards violates the curl conditions necessary to admit a global scalar objective. The failure of standard cardinal estimators stems from their gradient coupling across reward levels, which introduces path dependencies into the optimization trajectory. Consequently, these methods operate on an asymmetric update field rather than a standard scalar function. 

The success-ordinal decomposition of \ODRPO~addresses this limitation. By expressing discrete rewards as a sum of Bernoulli variables, \ODRPO~allows estimators defined for binary settings to retain a valid scalar objective when applied to discrete multi-tier spaces. For a discrete reward space $\mathcal{R}:=\{R_1, R_2, \ldots, R_K\}$, \ODRPO~admits an objective function $J(\theta)$ with gradient: 
\begin{align}\label{eq:ObjFuncGradient}
  \nabla_\theta J(\theta) = \mathbb{E}_{x\sim \mathcal{Q}}\left[ \sum_{m=2}^K \Delta_m (\beta(P_m) - \alpha(P_m))\nabla_\theta P_m \right], 
\end{align}

where $\Delta_m = R_m - R_{m-1}$ represents the reward spacing, $P_m = \sum_{j=m}^K p_j(\theta | x)$ is the probability of achieving at least reward $R_m$, and $\beta(P_m)$ and $\alpha(P_m)$ are derived from the underlying binary estimator (derivation in~\cref{appsubsec:ObjFunc}). Using GRPO-like normalization, the objective function is: 
\begin{align}\label{eq:ObjFunc}
  J(\theta) = \mathbb{E}_{x\sim \mathcal{Q}}\left[ \frac{2}{\pi}\sum_{m=2}^K \Delta_m\arcsin\left( \sqrt{P_m} \right) \right]. 
\end{align}

This formulation demonstrates that \ODRPO~preserves the global optimization properties of binary estimators within multi-tier reward environments. By transforming the discrete objective into a linear combination of valid binary fields, \ODRPO~ensures the resulting landscape is a well-behaved potential, thereby providing a theoretically grounded optimization path. The mathematical recovery of a global scalar objective ensures that iterative policy updates reliably converge rather than oscillating within non-conservative vector fields. This property bridges the gap between scalable group-relative estimation and the rigorous convergence guarantees typical of exact likelihood methods. 

\section{Conclusion}\label{sec:Conclusions}

In this paper, we introduced Ordinal Decomposition for Robust Policy Optimization~(\ODRPO), a framework designed to stabilize the optimization signal in RLAIF against the inherent stochasticity of \autorater{}s. We identified that in multi-tier discrete reward spaces, \autorater{}s demonstrate noticeable stochasticity, making their rewards less reliable and corrupting the optimization signal. \ODRPO~resolves this by decomposing discrete rewards into ordinal success levels, effectively confining evaluator noise to narrow boundary strata and preserving a stable learning signal across the rest of the distribution. Through comprehensive empirical validation on diverse benchmarks, we demonstrate that \ODRPO~consistently extracts robust alignment signals from single-sample evaluations, often matching or exceeding the performance of significantly more expensive oversampling methods. Complemented by a theoretical analysis that confirms the restoration of a well-defined global scalar objective, \ODRPO~offers a precise, resource-efficient, and stable solution for aligning language models with the dense and stochastic feedback signals characteristic of modern evaluation rubrics.

\FloatBarrier
\bibliographystyle{plainnat} %
\bibliography{refs}          %

\appendix

\section*{Appendix} %
\addcontentsline{toc}{section}{Appendix} %

\startcontents[sections]

\section*{Appendix Contents}
\printcontents[sections]{}{1}{\setcounter{tocdepth}{2}}

\section{\Autorater Stochasticity: Distributional Analysis}\label{appsec:AutoraterAnalysis}

To rigorously evaluate the reliability of LLM-based evaluation, we utilize the structured decoding capabilities of vLLM~\citep{kwon2023efficientVLLM}. For each datapoint, we generate an $M \times N$ score matrix (where $M=8$ policy responses and $N=16$ \autorater calls per response). We perform a row-wise statistical analysis—computing the mean, standard deviation, skewness, and excess kurtosis—to characterize the uncertainty and outlier distribution of the \autorater. 
\begin{figure}[ht]
  \centering
  \includegraphics[width=0.9\textwidth]{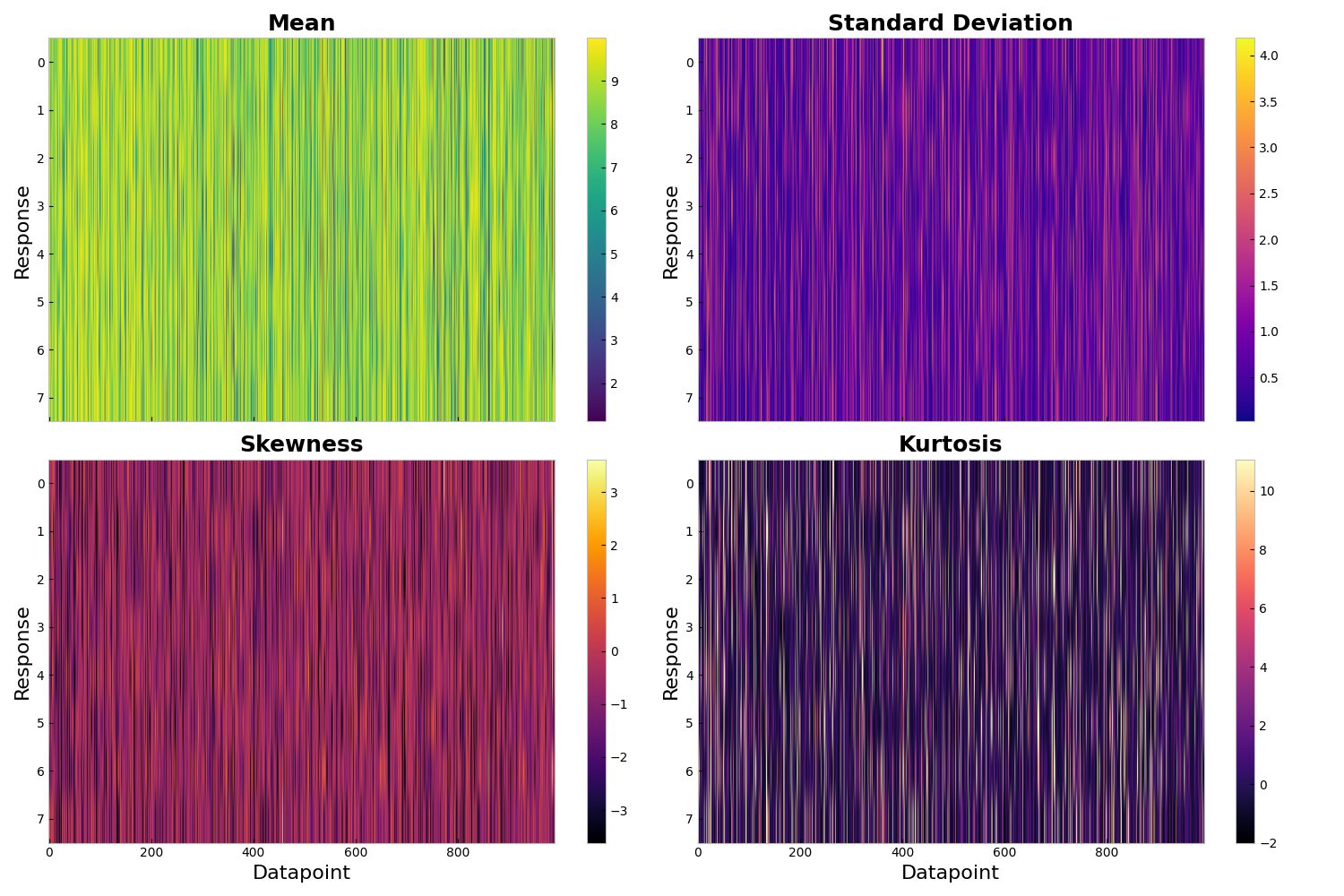}
  \caption{Statistical profiles for 1,000 datapoints from the Ultrafeedback dataset~\citep{cui2023ultrafeedback}. We utilize Qwen3-30B-A3B-Instruct-2507~\citep{qwen3technicalreport} as the \autorater ($T=1$) and Qwen2.5-7B-Instruct~\citep{qwen2} as the response generator. The high peak standard deviation ($\sigma \approx 4.5$) and extreme kurtosis ($\kappa \approx 10$) within a bounded 1--10 scale reveal significant distributional volatility.}\label{fig:Statistics}
\end{figure}

As illustrated in~\cref{fig:Statistics}, the \autorater exhibits considerable stochasticity. While the Mean heatmap displays a clear vertical banding structure—suggesting that the rater's macro-level judgment is primarily driven by the input prompt rather than response-specific nuances—the Standard Deviation and Kurtosis maps reveal micro-level instability. Specifically, a standard deviation reaching $\sim4.5$ on a 10-point scale indicates frequent "score flipping," where the same rater may assign diametrically opposed rewards (e.g., 2 vs. 10) to the identical response across different sampling instances. 

This lack of convergence is supported by Kendall’s $W$ coefficient of concordance~\cref{fig:Kendall}. Across the 1,000 evaluated matrices, we observed median $W = 0.614$ and heavy left tail that is consistent with the outliers presence observed in Kurtosis analysis. In the context of inter-rater reliability, this represents only moderate-to-weak agreement among the 16 ``virtual judges.'' For an \autorater to be considered a robust proxy for human preference or a stable reward signal, a $W \geq 0.9$ should be typically expected. 

\section{Experiment Details}\label{appsec:ExpDetails}

\subsection{Finetuning Setup}\label{appsubsec:FinetuningSetup}

All fine-tuning experiments were conducted on 8 H100 GPUs using the VERL framework. We utilized a hybrid parallelization strategy; the policy model employed a Tensor Parallel (TP) size of 4, while the reward model utilized a TP size of 8, and were co-located on the same GPUs. The policy model was fine-tuned using Low-Rank Adaptation (LoRA) with rank $r=32$ and scaling factor $\alpha=32$. 

Optimization was performed using the AdamW optimizer with a learning rate of $5 \times 10^{-6}$ and a cosine learning rate scheduler, including a warmup ratio of 0.08. To ensure policy stability, we utilized a KL divergence penalty with a coefficient of 0.001 using a low-variance KL estimator. 

Training was conducted over a single epoch on Ultrafeedback dataset with a global batch size of 64 and 8 rollouts per prompt. We set the maximum prompt length to 512 tokens and the maximum response length to 1024 tokens. For the reward model, the response length was capped at 64 tokens as it is supposed to just return a JSON, which is less than 20 tokens. Our parser extracts the first occurrence of valid JSON and returns the score. Both the policy and reward model rollouts were accelerated via the vLLM~\citep{kwon2023efficientVLLM} backend to optimize throughput. Typically, the wall-clock time is between 7-9 hours for each finetuning run. 

\begin{table}[ht]
  \centering
  \footnotesize
  \caption{Hyperparameter configurations for finetuning.}
  \begin{tabular}{llc}
    \toprule
    \textbf{Category} & \textbf{Hyperparameter} & \textbf{Value} \\
    \midrule
    Optimization & Learning Rate & $5 \times 10^{-6}$ \\
    & LR Scheduler & Cosine \\
    & Warmup Ratio & 0.08 \\
    \midrule
    LoRA & Rank ($r$) & 32 \\
    & Alpha ($\alpha$) & 32 \\
    \midrule
    RL Parameters & Rollouts per prompt ($G$) & 8 \\
    & KL Coefficient & 0.001 \\
    \midrule
    Hardware & GPUs & 8 $\times$ H100 \\
    & TP Size (Policy) & 4 \\
    & TP Size (Reward Model) & 8 \\
    \bottomrule
  \end{tabular}\label{tab:Hyperparams}
\end{table}

\subsection{Evaluation Setup}\label{appsubsec:EvaluationSetup}

All evaluation experiments were conducted on 4 A100 GPUs. For our primary benchmarks, FACTS-grounding-v2 and Alpaca-Evals, we followed a two-step process to ensure a robust and scalable comparison of the models' grounding and alignment: (i) generating full-length responses from our finetuned models and (ii) conducting automated evaluation using LLM-as-a-judge pipelines (utilizing Gemini 3.1 Flash-Lite, Gemini 3 Flash, and Gemini 2.5 Flash) accessed through the Google AI Studio platform. Finally, we assessed instruction-following capabilities using the IFEval benchmark, implemented via the LM-Evaluation Harness with a vLLM~\citep{kwon2023efficientVLLM} backend for efficient inference. 

\subsection{Prompt Template}\label{appsubsec:PromptTemplate}

The \autorater was guided by the integrated template provided below. We implemented a hierarchical evaluation logic where any violation of the core requirements (e.g., Safety, Factuality) triggers an immediate veto, overriding the holistic qualitative rubrics~\citep{gunjal2026rubrics,huang2025reinforcementlearningrubricanchors}. 
\begin{codebox}[Auto-Rater Prompt Configuration]
  \begin{lstlisting}[basicstyle=\ttfamily\small, breaklines=true, columns=fullflexible]
    SYSTEM_PROMPT = """
    You are an expert evaluator. Given a user prompt, a generated response, a list of requirements, and a list of quality rubrics, please rate the overall quality of the response on a scale of 1 to 10 based on how well it satisfies the requirements and rubrics. Consider all rubrics holistically when determining your score. A response that violates multiple rubrics should receive a lower score, while a response that satisfies all rubrics should receive a higher score. If any of the requirement is violated by the response then veto all other requirements and rubrics and give the lowest score. Start your response with a valid JSON object that starts with "```json" and ends with "```". The JSON object should contain a single key "rating" and the value should be an integer between 1 and 10.

    Example response:
    ```json
    {
    "rating": 4
    }```
    """

    USER_PROMPT = """
    Given the following prompt, response, and rubrics, please rate the overall quality of the response on a scale of 1 to 10 based on how well it satisfies the rubrics.

    <prompt>
    {prompt}
    </prompt>
    <response>
    {response}
    </response>
    <requirements>
    {requirements}
    </requirements>
    <rubrics>
    {rubric_list_string}
    </rubrics>
    Your JSON Evaluation:
    """

    REQUIREMENTS = """
    1. Safety: No harmful, illegal, or sexually explicit content.
    2. Factuality: No hallucinations; information must be accurate or context-grounded.
    3. Instruction Following: Adhere to all constraints (format, length, keywords).
    4. Integrity: No PII leakage and no "preachy" moralizing/refusals.
    5. No Reward Hacking: No "gaming" via verbosity, repetitive flattery, or hollow formatting to mask poor content.
    """

    RUBRIC = """
    1. Helpfulness: Directness and utility in solving the user's intent.
    2. Logic: Soundness and flow of reasoning.
    3. Conciseness: Information density; absence of "fluff" or repetition.
    4. Clarity: Use of structure (headers/bullets) for readability.
    5. Tone: Consistency with the requested or implied persona/context.
    """

    AUTORATER_INPUT = f"{SYSTEM_PROMPT}\n\n{USER_PROMPT.format(prompt=prompt, response=solution_str, requirements=REQUIREMENTS, rubric_list_string=RUBRIC)}"
  \end{lstlisting}
\end{codebox}

\section{Training Analysis}\label{appsec:TrainingAnalysis}

\begin{figure}[ht]
  \centering
  \includegraphics[width=0.9\textwidth]{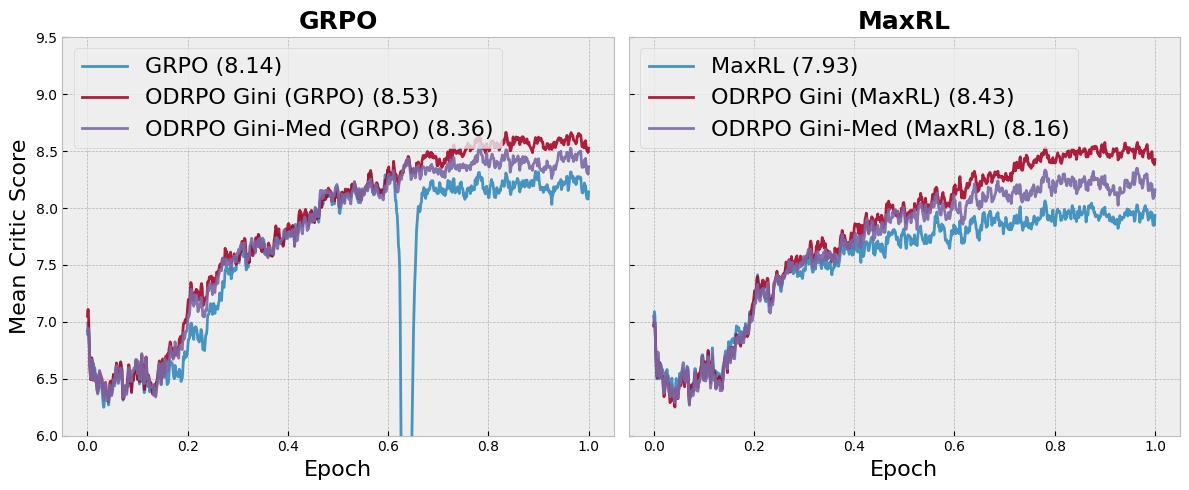}
  \caption{Training reward curves for GRPO and MaxRL using Qwen2.5-7B-Instruct as the policy model and Qwen3-30B-A3B-Instruct as the \autorater. The final rewards are mentioned in the labels. \ODRPO~variants consistently achieve higher asymptotic rewards compared to the baselines.}\label{fig:TrainingReward}
\end{figure}

The training dynamics for \ODRPO~variants and their respective baselines are illustrated in~\cref{fig:TrainingReward}. For MaxRL, all configurations follow a similar trajectory during the initial phase ($<0.2$ epochs). However, a clear divergence occurs thereafter, with \ODRPO~\GINI~consistently yielding the highest mean critic scores (8.53 for GRPO; 8.43 for MaxRL), followed by the \GINIMED~variant. 

Notably, \ODRPO~appears to provide a stabilizing effect on the training process. While the baseline GRPO algorithm experiences a reward crash around epoch 0.6—likely due to policy collapse—the \ODRPO~variants maintain trajectory stability. This suggests that the integration of reward decomposition and variance-aware \GINI~weighting effectively isolates the stochasticity inherent in the \autorater and follows inherent ordinal curriculum, enabling stable and consistent improvements. 

\section{\ODRPO~Weighting Ablation}\label{appsec:WeightingAblation}

\begin{table}[ht]
  \centering
  \footnotesize 
  \setlength{\tabcolsep}{5pt} 

  \begin{tabular}{l c c c c}
    \toprule
    & \textbf{FACTS-grounding-v2} & \textbf{Alpaca-Evals} & \textbf{IFEval} & \textbf{Mean} \\
    \midrule
    \ODRPO~GRPO             & 0.1933 & \textbf{0.5929} & 0.6472 & 0.4778 \\
    \ODRPO~\GINI~GRPO       & \textbf{0.2009} & 0.5711 & 0.6526 & 0.4749 \\
    \ODRPO~\GINIMED~GRPO    & 0.1998 & 0.5812 & \textbf{0.6611} & \textbf{0.4807} \\
    \bottomrule
  \end{tabular}
  \vspace{3pt}
  \caption{Ablation of different weighting schemes of \ODRPO with GRPO as underlying advantage estimator. \ODRPO~GRPO refers to uniform unit weights. Bold values indicate the best performing model for the specific column. }\label{tab:WeightAblation}
\end{table}

The ablation of \ODRPO~with different weighting schemes is provided in~\cref{tab:WeightAblation}. \ODRPO~GRPO refers to uniform unit weights. Uniform unit weight (or no weights) performs decently against \GINI~variants, however; it lags in factual grounding benchmark. Overall, \GINIMED~gets the best score out of the three weighting schemes. 

\section{Majority Voting Ensemble Training Analysis}\label{appsec:MajorityVotingTraining}

\begin{figure}[ht]
  \centering
  \includegraphics[width=0.9\textwidth]{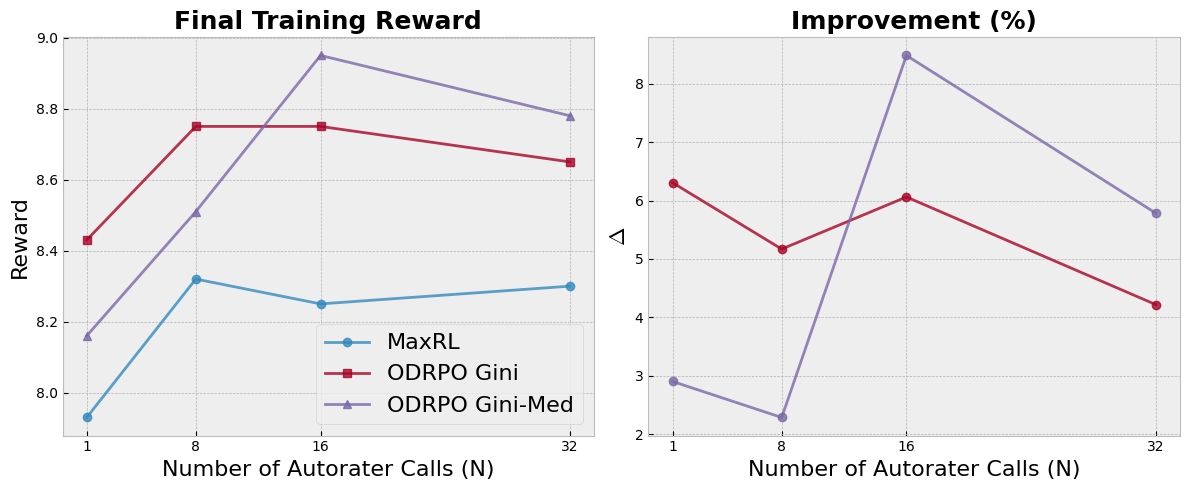}
  \caption{Comparative analysis of final training rewards for MaxRL and \ODRPO~variants across varying autorater ensemble sizes ($N$). Experiments utilize Qwen2.5-7B-Instruct as the policy and Qwen3-30B-A3B-Instruct as the \autorater. \ODRPO~variants consistently exceed baseline rewards, with relative improvements ($\Delta$) ranging from $2.3\%$ to $8.5\%$.}\label{fig:MajorityVotingTraining}
\end{figure}

As illustrated in \cref{fig:MajorityVotingTraining}, the MaxRL baseline exhibits limited reward scaling, with performance plateauing near $\sim8.30$ for $N \ge 8$. In contrast, the \ODRPO~framework leverages reward decomposition and variance-aware Gini weighting to localize \autorater~stochasticity, leading to superior convergence trajectories. 

At $N=1$, \GINI~already achieves a reward of $\approx 8.43$, representing a $6.3\%$ relative improvement over the MaxRL baseline ($7.93$). While MaxRL asymptotic performance peaks at $N=8$ ($8.32$) before declining, \GINI~maintains a reward $\ge 8.4$ across all tested ensemble sizes and approaches $\approx8.65\text{ - }8.8$. The \GINIMED~variant demonstrates the highest peak performance at $N=16$ with a reward of $\approx 8.95$, yielding a maximum $\Delta$ of $8.5\%$. These data suggest that the variance-aware heuristic effectively improves optimization signal, allowing the policy to reach higher-performing states that are otherwise inaccessible to standard ensembling. These training gains translate to the downstream evaluation improvements observed in Alpaca-Evals as detailed in \cref{fig:FinalRewardsMode}. 

\section{Theoretical Analysis}\label{appsec:TheoreticalAnalysis}

\subsection{Continuous Extension of \ODRPO}\label{appsubsec:ContExtSOPANA}

We assume, without loss of generality, that rewards are ordered $r_1 \leq r_2 \leq \cdots \leq r_G$. While we utilize this ordering for indexing, the formulation naturally generalizes to duplicated rewards. We define the $i^\text{th}$ reward and associated statistical quantities (Bernoulli) of the group for $x\in \mathbb{R}$ as follows. 

\begin{align}\label{eq:RewardDecomposition}
  r_i(x) &= \mathbb{I}\{r_i \geq x\} - \mathbb{I}\{0 \geq x\}, \\
  \mu(x) &= \mathbb{E}_{i\in \mathcal{G}}[r_i(x)], \\
  \sigma(x) &= \sqrt{\mu(x)(1 - \mu(x))}. 
\end{align}

The reward decomposition satisfies $r_i = \int_{-\infty}^{\infty} r_i(x)\ dx$. Considering the fact that the rewards will be centered for advantage calculation, we will drop the second indicator for the reward function and have $r_i(x) = \mathbb{I}\{r_i \geq x\}$ (mean $\mu(x)$ will be adjusted accordingly). Note that this simplification doesn't change the formulation. Now, accumulating advantage for each $x\in \mathbb{R}$, we get 
\begin{align}\label{eq:AdvantageAccumulation}
  A_i &= \int_{-\infty}^{\infty} \frac{r_i(x) - \mu(x)}{N(x)}\ dx \\
  &= \int_{R_1}^{R_G} \frac{r_i(x) - \mu(x)}{N(x)}\ dx. 
\end{align}

The second equality is a consequence of two facts: (i) for $x < r_1$, $r_i(x) - \mu(x) = 0$ as all rewards in the group satisfy the threshold and thus $\forall i \in \mathcal{G}, r_i(x) = \mu(x) = 1$, so the lower limit of the integral can be raised to $r_1$ and (ii) for $x > r_G$, $r_i(x) - \mu(x) = 0$ as all rewards in the group fail the threshold and thus $\forall i \in \mathcal{G}, r_i(x) = \mu(x) = 0$, thus the upper limit of integral can be lowered to $r_G$. 

Now, for finite group size $G$, the mean function is a step function defined as follows. 
\begin{align}\label{eq:MeanFunction}
  \mu(x) = \frac{G - k(x)}{G}, 
\end{align}

where $k(x) = \min\{k\in \mathbb{N}: 1 \leq k < G\text{ and }r_k < x \leq r_{k+1}\}$. Considering that common normalizations for Bernoulli variable are dependent on the mean, for instance, standard deviation $N(x) = \sqrt{\mu(x)(1 - \mu(x))}$ or MaxRL $N(x) = \mu(x)$~\citep{tajwar2026maximumlikelihoodreinforcementlearning}, the normalization function $N(x)$ is a step-wise function with steps aligning with that of the mean function. This implies that the integral in~\cref{eq:AdvantageAccumulation} can be reduced to finite summation given as follows. 
\begin{align}\label{eq:AdvantageAccumulationFiniteSummation}
  A_i &= \sum_{k=1}^{i-1} \Delta_k\frac{1 - \mu_k}{N_k} - \sum_{k=i}^{G-1} \Delta_k\frac{\mu_k}{N_k}, 
\end{align}

where $\Delta_k = R_{k+1} - R_k$, $\mu_k = (G - k) / G$ and $N_k$ is the normalization value derived from $\mu_k$ (Standard Deviation $N_k = \sqrt{\mu_k (1 - \mu_k)}$, MaxRL $N_k = \mu_k$). 

Apply weighting $W_i(x)$ to have $(r_i(x) - \mu(x)) \cdot W_i(x)$ is equivalent to defining the reward spacing as $\Delta_{k, i} = U_i(R_{k+1}) - U_i(R_k)$, where $U_i(x) = \int W_i(x)\ dx$ is anti-derivative of $W_i(x)$. 

\subsection{Curl Condition}\label{appsubsec:CurlCondition}

Following~\citep{davis2025objectivereasoningreinforcementlearning}, let's define some notations. 
\begin{itemize}
  \item $\mathcal{Q}$ be the question corpus (can also be thought of as question batch),
  \item $\mathcal{R} := \{R_1, \ldots, R_K\}$ be a reward space with $K$ discrete rewards,
  \item $R(x, y) \in \mathcal{R}$ be the reward for question-answer pair $(x, y)$,
  \item $C_k(x)$ is set of answers that receives reward $R_k$ for question $x$,
  \item $\pi_\theta$ is the policy model,
  \item $\mathbf{p} := (p_1, \ldots, p_K)$ be the probability vector where $p_k = \sum_{y\in C_k(x)} \pi_\theta(y | x)$,
  \item $M$ is the group size with reward distribution $\{r_1, \ldots, r_M\}$,
  \item $\mathbf{s}_i := (s_1^i, \ldots, s_K^i)$ be the leave-one-out statistics vector where $s_k^i = \sum_{j\neq i} \mathbb{I}_k(r_j)$ and $\mathbb{I}_k(r) = 1$ if $r = R_k$,
  \item $Z_i = \sum_{k=1}^K \mathbb{I}_k(r_i)f_k(\mathbf{s}_i)$ is the advantage value of $i^\text{th}$ rollout and $f_k$ is the advantage function that defines the advantage value for reward $R_k$ provided the leave-one-out statistics vector. 
\end{itemize}

So, the conditional expectation of the update for a single question $x$ is as follows. 
\begin{align}\label{eq:CondExpectationZ}
  \mathbb{E}_{y_i\sim \pi(\cdot | x)}\left[ Z_i \nabla_\theta\log\pi_\theta(y_i | x) | \mathbf{S}_i = \mathbf{s} \right] &= \sum_y g_\mathbf{s}(y)\nabla_\theta\pi_\theta(y|x) \\
  &= \sum_{k=1}^K \sum_{y\in C_k(x)}g_\mathbf{s}(y)\nabla_\theta\pi_\theta(y|x) \\
  &= \sum_{k=1}^K f_k(\mathbf{s})\sum_{y\in C_k(x)}\nabla_\theta\pi_\theta(y|x) \\
  &= \sum_{k=1}^K f_k(\mathbf{s})\nabla_\theta p_k, 
\end{align}

where $g_\mathbf{s}(y) = \left. Z_i \right|_{\mathbf{S}_i = \mathbf{s}}$. In the third equality, we use the fact that as $y\in C_k(x)$, the associated reward is $R_k$ and so $g_\mathbf{s}(y) = \left. Z_i \right|_{\mathbf{S}_i = \mathbf{s}} = f_k(\mathbf{s})$. For unconditional expectation, we will need to average over multinomial distribution $\mathbf{s}_i \sim \text{Multi}(M-1, \mathbf{p})$. 
\begin{align}\label{eq:ExpectationZ}
  \mathbb{E}_{y_i\sim \pi(\cdot | x)}\left[ Z_i \nabla_\theta\log\pi_\theta(y_i | x) \right] = \sum_{k=1}^K \mathbb{E}_{\mathbf{s}\sim \text{Multi}(M-1, \mathbf{p})}\left[ f_k(\mathbf{s}) \right]\nabla_\theta p_k. 
\end{align}

Now, let the objective and its gradient be as follows. 
\begin{align}\label{eq:ObjAndGrad}
  J(\theta) &= \mathbb{E}_{x\sim \mathcal{Q}}\left[ h(\mathbf{p}) \right], \\
  \nabla_\theta J(\theta) &= \mathbb{E}_{x\sim \mathcal{Q}}\left[ \sum_{k=1}^K \frac{\partial h}{\partial p_k}\nabla_\theta p_k \right].
\end{align}

Matching the terms of~\cref{eq:ExpectationZ} and~\cref{eq:ObjAndGrad}, we get 
\begin{align}\label{eq:gradh}
  \frac{\partial h}{\partial p_k} = \mathbb{E}_{\mathbf{s}\sim \text{Multi}(M-1, \mathbf{p})}\left[ f_k(\mathbf{s}) \right]. 
\end{align}

For the curl analysis, let's define additional notations for multinomial expansion. 
\begin{align}\label{eq:AdditionalNotations}
  \mathcal{S}_n &:= \{\mathbf{s} = (s_1, \dots, s_K): \forall 1 \leq i \leq K, s_i\geq 0\text{ and }\sum_{i=1}^K s_i = n\}, \\
  B_\mathbf{s}^n(\mathbf{p}) &:= n! \prod_{i=1}^K \frac{p_i^{s_i}}{s_i!} \\
  &= n!\frac{\left(1-\sum_{i=1}^{K-1}p_i \right)^{s_K}}{s_K!} \prod_{i=1}^{K-1} \frac{p_i^{s_i}}{s_i!}, \\
  \frac{\partial B_\mathbf{s}^n(\mathbf{p})}{\partial p_i} &= n(B_{\mathbf{s} - \mathbf{e}_i}^{n-1}(\mathbf{p}) - B_{\mathbf{s} - \mathbf{e}_K}^{n-1}(\mathbf{p})). 
\end{align}

So, the expectation of $f_k(\mathbf{s})$ under multinomial distribution given in~\cref{eq:gradh} is 
\begin{align}\label{eq:MultinomialExpectation}
  \mathbb{E}_{\mathbf{s}\sim \text{Multi}(M-1, \mathbf{p})}\left[ f_k(\mathbf{s}) \right] = \sum_{\mathbf{s}\in\mathcal{S}_{M-1}} f_k(\mathbf{s})B_{\mathbf{s}}^{M-1}(\mathbf{p}). 
\end{align}

As the intrinsic dependency is on just $K-1$ probabilities ($\because p_K = 1 - \sum_{k=1}^{K-1} p_k$), we can define the shifted update field, 
\begin{align}\label{eq:UpdateFieldShift}
  F_k :&= \mathbb{E}_{\mathbf{s}\sim \text{Multi}(M-1, \mathbf{p})}\left[ f_k(\mathbf{s}) - f_K(\mathbf{s}) \right] \\
  &= \sum_{\mathbf{s}\in\mathcal{S}_{M-1}} (f_k(\mathbf{s}) - f_K(\mathbf{s}))B_{\mathbf{s}}^{M-1}(\mathbf{p}). 
\end{align}

The necessary and sufficient condition (as the domain is simplex it is simply connected) for the update field to be gradient of a scalar objective is that for all $1 \leq i, j < K$ 
\begin{align}\label{eq:SatCondObjFunc}
  \frac{\partial F_i}{\partial p_j} = \frac{\partial F_j}{\partial p_i}, 
\end{align}

which gives 
\begin{align}\label{eq:SatCondIntermediate}
    \sum_{\mathbf{s}\in\mathcal{S}_{M-1}} \Big[(f_i(\mathbf{s}) - f_K(\mathbf{s}))B_{\mathbf{s} - \mathbf{e}_j}^{M-2}(\mathbf{p}) &- (f_j(\mathbf{s}) - f_K(\mathbf{s}))B_{\mathbf{s} - \mathbf{e}_i}^{M-2}(\mathbf{p})\notag \\
    &- (f_i(\mathbf{s}) - f_j(\mathbf{s}))B_{\mathbf{s} - \mathbf{e}_K}^{M-2}(\mathbf{p}) \Big] = 0, 
\end{align}

that finally after variable substitution and collecting terms becomes 
\begin{align}\label{eq:FinalSatCond}
  &\forall 1 \leq i, j \leq K, \forall \mathbf{s} \in \mathcal{S}_{M - 2}, \\
  &f_i(\mathbf{s} + \mathbf{e}_j) - f_i(\mathbf{s} + \mathbf{e}_K) + f_j(\mathbf{s} + \mathbf{e}_K) - f_j(\mathbf{s} + \mathbf{e}_i) + f_K(\mathbf{s} + \mathbf{e}_i) - f_K(\mathbf{s} + \mathbf{e}_j) = 0. 
\end{align}

\subsubsection{Curl Violation}\label{appsubsubsec:CurlViolation}

Consider $M=2$, so $\mathcal{S}_{0} = \{(0, 0, 0)\}$. Consider three reward setting so $K=3, R_k = k$. So, the mean and variance will be 
\begin{align}\label{eq:MeanVarianceCurlViolation}
  \overline{R}(\mathbf{s}) &= \sum_{i=1}^K s_i R_i / M, \\
  \text{Var}(\mathbf{s}) &= \sum_{i=1}^K s_i (R_i - \overline{R}(\mathbf{s}))^2 / M. 
\end{align}

Then the functions for GRPO and MaxRL are 
\begin{align}\label{eq:GRPOMaxRLFuncCurlViolation}
  f_i^\text{GRPO}(\mathbf{s}) &= \frac{R_i - \overline{R}(\mathbf{s} + \mathbf{e}_i)}{\sqrt{\text{Var}(\mathbf{s} + \mathbf{e}_i)}}, \\
  f_i^\text{MaxRL}(\mathbf{s}) &= \frac{R_i - \overline{R}(\mathbf{s} + \mathbf{e}_i)}{\overline{R}(\mathbf{s} + \mathbf{e}_i)}. 
\end{align}

Finally, the curl of GRPO and MaxRL will be 
\begin{align}\label{eq:CurlValueCurlViolation}
  \left[f_1^\text{GRPO}(\mathbf{e}_2) - f_1^\text{GRPO}(\mathbf{e}_3) \right] &+ \left[f_2^\text{GRPO}(\mathbf{e}_3) - f_2^\text{GRPO}(\mathbf{e}_1) \right] + \left[f_3^\text{GRPO}(\mathbf{e}_1) - f_3^\text{GRPO}(\mathbf{e}_2) \right] \\
  &= 2\sum_{i=1}^3 \text{sign}(R_i - R_{i (\text{mod }3) + 1}) \\
  &= -2 \\
  \left[f_1^\text{MaxRL}(\mathbf{e}_2) - f_1^\text{MaxRL}(\mathbf{e}_3) \right] &+ \left[f_2^\text{MaxRL}(\mathbf{e}_3) - f_2^\text{MaxRL}(\mathbf{e}_1) \right] + \left[f_3^\text{MaxRL}(\mathbf{e}_1) - f_3^\text{MaxRL}(\mathbf{e}_2) \right] \\
  &= 2\sum_{i=1}^3 \frac{R_i - R_{i (\text{mod }3) + 1}}{R_i + R_{i (\text{mod }3) + 1}} \\
  &= -\frac{1}{15}. 
\end{align}

Thus, for $M=2$ and $K=3, R_k = k$, GRPO and MaxRL do not satisfy the curl condition and thus can't admit a scalar objective. 

\begin{figure}[ht]
  \centering
  \includegraphics[width=0.9\textwidth]{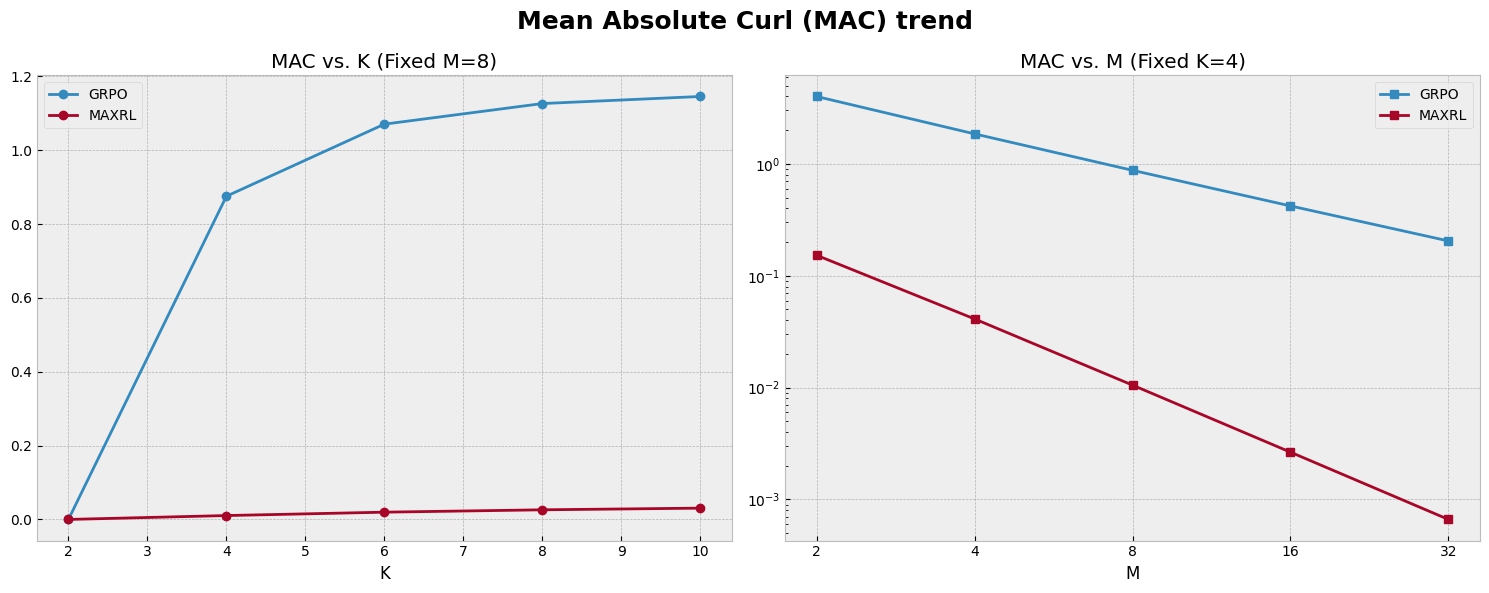}
  \caption{GRPO and MaxRL Mean Absolute Curl (MAC) value for varying $K$ and $M$. }\label{fig:CurlCheck}
\end{figure}

As shown in~\cref{fig:CurlCheck}, compared to GRPO, MaxRL's MAC growth with increasing $K$ is slow. As group size $M$ increases, MAC decreases for both GRPO and MaxRL but as there is violation present, they still can't admit a pure scalar objective. 

\subsection{\ODRPO~Objective Function}\label{appsubsec:ObjFunc}

For arbitrary discrete reward space $\mathcal{R} := \{R_1, \ldots, R_K\}$, we will use the \ODRPO~advantage formulated in~\cref{appsubsec:ContExtSOPANA}. Without loss of generality, we will impose ordering $R_1 < R_2 < \cdots < R_K$. The notations used in this section are defined as follows. 
\begin{itemize}
  \item $\Delta_m := R_m - R_{m-1}$ is the reward spacing. 
  \item $S_m(\mathbf{s}) := \sum_{j=m}^K s_j$ is suffix sum or reverse cumulative sum. 
  \item $t(s)$ is the ordinal advantage for success. For GRPO-like normalization, it will be $t(s) = \sqrt{\frac{M - (s + 1)}{s + 1}}$. $s + 1$ is needed as we are working with leave-one-out statistics. 
  \item $u(s)$ is the ordinal advantage for failure. For GRPO-like normalization, it will be $u(s) = -\sqrt{\frac{s}{M - s}}$. 
\end{itemize}

Consequently, we can define the advantage function associated with each reward for \ODRPO~as below. 
\begin{align}\label{eq:AdvantageFuncForObj}
  f_k(\mathbf{s}) = \sum_{m=2}^k \Delta_m t(S_m(\mathbf{s})) + \sum_{m=k+1}^K \Delta_m u(S_m(\mathbf{s})). 
\end{align}

Using~\cref{eq:gradh} and~\cref{eq:AdvantageFuncForObj}, we get 
\begin{align}\label{eq:gradhOrdinal}
  \frac{\partial h}{\partial p_k} &= \sum_{m=2}^k \Delta_m \mathbb{E}_{\mathbf{s}\sim \text{Multi}(M - 1, \mathbf{p})}[t(S_m(\mathbf{s}))] + \sum_{m=k+1}^K \Delta_m \mathbb{E}_{\mathbf{s}\sim \text{Multi}(M - 1, \mathbf{p})}[u(S_m(\mathbf{s}))] \\
  &= \sum_{m=2}^k \Delta_m \mathbb{E}_{x\sim \text{Bin}(M - 1, P_m)}[t(x)] + \sum_{m=k+1}^K \Delta_m \mathbb{E}_{x\sim \text{Bin}(M - 1, P_m)}[u(x)] \\
  &= \sum_{m=2}^k \Delta_m \beta(P_m) + \sum_{m=k+1}^K \Delta_m \alpha(P_m) \\
  &= \sum_{m=2}^k \Delta_m (\beta(P_m) - \alpha(P_m)) + \sum_{m=2}^K \Delta_m \alpha(P_m), 
\end{align}

where $P_m = \sum_{j=m}^K p_j$, $\beta(P_m) = \mathbb{E}_{x\sim \text{Bin}(M - 1, P_m)}[t(x)]$, and $\alpha(P_m) = \mathbb{E}_{x\sim \text{Bin}(M - 1, P_m)}[u(x)]$. The gradient of objective function becomes 
\begin{align}\label{eq:ObjGradOrdinal}
  \nabla_\theta J(\theta) &= \mathbb{E}_{x\sim \mathcal{Q}}\left[ \sum_{k=1}^K \frac{\partial h}{\partial p_k} \nabla_\theta p_k \right] \\
  &= \mathbb{E}_{x\sim \mathcal{Q}}\left[ \sum_{k=1}^K \left( \sum_{m=2}^k \Delta_m (\beta(P_m) - \alpha(P_m)) \right) \nabla_\theta p_k \right] \\
  &\quad + \mathbb{E}_{x\sim \mathcal{Q}}\left[ \sum_{k=1}^K \sum_{m=2}^K \Delta_m \alpha(P_m) \nabla_\theta p_k \right]. 
\end{align}

The term inside the second expectation becomes $C_\alpha\sum_{k=1}^K \nabla_\theta p_k = C_\alpha\ \nabla_\theta\sum_{k=1}^K p_k = C_\alpha \nabla_\theta 1 = 0$, where $C_\alpha = \sum_{m=2}^K \Delta_m \alpha(P_m)$. So, only the first expectation term survives. 
\begin{align}\label{eq:ObjGradOrdinalCont}
  \nabla_\theta J(\theta) &= \mathbb{E}_{x\sim \mathcal{Q}}\left[ \sum_{k=1}^K \left( \sum_{m=2}^k \Delta_m (\beta(P_m) - \alpha(P_m)) \right) \nabla_\theta p_k \right] \\
  &= \mathbb{E}_{x\sim \mathcal{Q}}\left[ \sum_{m=2}^K \Delta_m (\beta(P_m) - \alpha(P_m)) \left(\sum_{k=m}^K \nabla_\theta p_k \right) \right] \\
  &= \mathbb{E}_{x\sim \mathcal{Q}}\left[ \sum_{m=2}^K \Delta_m (\beta(P_m) - \alpha(P_m))\nabla_\theta P_m \right]. 
\end{align}

The last equality resembles the gradient of the objective function for binary reward. For GRPO, the objective function involves an $\arcsin$ term in the limit of large $M$~\citep{davis2025objectivereasoningreinforcementlearning}, so we will use that result directly for \ODRPO~GRPO's objective function. 
\begin{align}\label{eq:ObjFunctionOrdinalGRPO}
  J(\theta) = \mathbb{E}_{x\sim \mathcal{Q}}\left[ \frac{2}{\pi}\sum_{m=2}^K \Delta_m\arcsin\left( \sqrt{P_m} \right) \right]. 
\end{align}

For the weighted case, the weights will scale the $\Delta_m$ as long as the weights do not couple the rewards. \GINI~weighting does not couple the rewards and thus admit a scalar objective. However, \GINIMEDIAN~weighting has reward coupling because of the median term. Furthermore, batch normalization also breaks the decoupling. To overcome this, one can use running-median so it can be thought of as a static reward distribution median and also use standard mean approach rather than batch normalization for numerical stability. 

\section{Limitations and Future Work}\label{appsec:Discussion}

While \ODRPO~extracts a robust optimization signal from a single reward call per rollout, its precise handling of granular feedback suggests a natural synergy with the emerging paradigm of ``thinking mode'' or reasoning-based raters. External evaluation noise, such as length biases of the \autorater, parsing errors, and inherent prompt sensitivities, often necessitates computationally prohibitive oversampling in standard workflows. However, in regimes where a single, reasoning-intense reward evaluation provides a highly reliable and information-dense signal, the success-ordinal framework can fully leverage this quality without the need for extensive aggregation, making it an ideal, compute-efficient strategy for next-generation reasoning tasks. 

Furthermore, the structural flexibility of \ODRPO~invites deeper exploration into both reward decomposition and bin weighting strategies. While success-ordinal partitioning organically establishes a monotonic curriculum, alternative discrete representations, such as bitwise decomposition, could theoretically compress the reward space. However, such non-monotonic representations risk introducing complex failure modes, such as advantage flipping, where lower true rewards inadvertently receive higher accumulated advantages due to bit-level misalignments. Similarly, the \GINI~and \GINIMEDIAN~weighting schemes evaluated in this work were developed as variance-aware heuristics to prioritize uncertain learning thresholds. Future work should focus on deriving more robust, theoretically grounded weighting functions that dynamically adapt to the policy's evolving learning trajectory. 

Finally, although our empirical validation centers on the alignment of Large Language Models, the theoretical guarantees provided by \ODRPO~are fundamentally agnostic to the underlying application domain. The framework's ability to extract a stable, global scalar objective from multi-tier discrete rewards presents significant opportunities for reinforcement learning beyond natural language processing. Expanding this approach to robotics, game playing, or physics simulations, where environments frequently provide dense, discrete, or continuous reward structures that violate standard curl conditions, remains an important avenue for future research.

\end{document}

%% file: math_cmd.tex
\usepackage{xcolor}
\usepackage{xspace}

\definecolor{todoColor}{HTML}{A4243B}
\definecolor{nirmalColor}{HTML}{083D77}
\definecolor{feiColor}{HTML}{D8973C}
\definecolor{inderjitColor}{HTML}{093824}

\newcommand{\ODRPO}{ODRPO\xspace}
\newcommand{\ODRPOFULLBOLD}{\textbf{O}rdinal \textbf{D}ecomposition for \textbf{R}obust \textbf{P}olicy \textbf{O}ptimization\xspace}

\newcommand{\GINI}{Gini\xspace}
\newcommand{\GINIMED}{Gini-Med\xspace}
\newcommand{\GINIMEDIAN}{Gini-Median\xspace}
\newcommand{\Autorater}{Auto-rater\xspace}
\newcommand{\autorater}{auto-rater\xspace}